\documentclass[letterpaper]{article} 
\usepackage{aaai2026}  
\usepackage{times}  
\usepackage{helvet}  
\usepackage{courier}  
\usepackage[hyphens]{url}  
\usepackage{graphicx} 
\usepackage{natbib}  
\usepackage{caption} 
\usepackage{algorithm}
\usepackage{algorithmic}

\usepackage{newfloat}
\usepackage{listings}
\DeclareCaptionStyle{ruled}{labelfont=normalfont,labelsep=colon,strut=off} 
\floatstyle{ruled}
\newfloat{listing}{tb}{lst}{}
\floatname{listing}{Listing}
\title{Bidirectional Channel-selective Semantic Interaction for Semi-Supervised Medical Segmentation}
\author{
    Kaiwen Huang\textsuperscript{\rm 1}, 
    Yizhe Zhang\textsuperscript{\rm 1},
    Yi Zhou\textsuperscript{\rm 2},
    Tianyang Xu\textsuperscript{\rm 3},
    Tao Zhou\textsuperscript{\rm 1}\thanks{Corresponding author.}
}
\affiliations{

    \textsuperscript{\rm 1}School of Computer Science and Engineering,
    Nanjing University of Science and Technology, China.\\
    \textsuperscript{\rm 2}School of Computer Science and Engineering, Southeast University, China.\\
    \textsuperscript{\rm 3}School of Artificial Intelligence and Computer Science, Jiangnan University, China.\\
    taozhou.ai@gmail.com
}

\usepackage{bibentry}

\usepackage{overpic}
\usepackage{amsmath}
\usepackage{amssymb}
\usepackage{multirow}
\usepackage{rotating}
\usepackage{booktabs}

\usepackage{graphicx,verbatim}

\usepackage[table]{xcolor}
\usepackage{marvosym} 
\usepackage{overpic}

\def\eg{\emph{e.g.}}

\begin{document}

\maketitle

\begin{abstract}

Semi-supervised medical image segmentation is an effective method for addressing scenarios with limited labeled data. Existing methods mainly rely on frameworks such as mean teacher and dual-stream consistency learning. These approaches often face issues like error accumulation and model structural complexity, while also neglecting the interaction between labeled and unlabeled data streams. To overcome these challenges, we propose a Bidirectional Channel-selective Semantic Interaction~(BCSI) framework for semi-supervised medical image segmentation. 
First, we propose a Semantic-Spatial Perturbation~(SSP) mechanism, which disturbs the data using two strong augmentation operations and leverages unsupervised learning with pseudo-labels from weak augmentations. Additionally, we employ consistency on the predictions from the two strong augmentations to further improve model stability and robustness. Second, to reduce noise during the interaction between labeled and unlabeled data, we propose a Channel-selective Router~(CR) component, which dynamically selects the most relevant channels for information exchange. This mechanism ensures that only highly relevant features are activated, minimizing unnecessary interference. Finally, the Bidirectional Channel-wise Interaction~(BCI) strategy is employed to supplement additional semantic information and enhance the representation of important channels. 
Experimental results on multiple benchmarking 3D medical datasets demonstrate that the proposed method outperforms existing semi-supervised approaches. 

\end{abstract}

\begin{links}
\link{Code}{https://github.com/taozh2017/BCSI}
\end{links}

\section{Introduction}

Medical image segmentation plays a crucial role in clinical diagnosis, enabling precise identification of lesion areas, organs, and other anatomical structures~\cite{jiao2024learning, mei2025survey}. However, current models~\cite{cao2022swin, zhou2023nnformer} rely heavily on large annotated datasets for training. This is particularly challenging for medical images, as annotating them requires expert knowledge and significant manual effort. To mitigate this issue, Semi-Supervised Learning (SSL) methods~\cite{chen2021semi, tarvainen2017mean, yao2022enhancing,zhang2025s} leverage limited labeled data alongside abundant unlabeled data, reducing annotation costs while enhancing model performance.

\begin{figure}[t]
\centering
\includegraphics[width=0.99\columnwidth]
{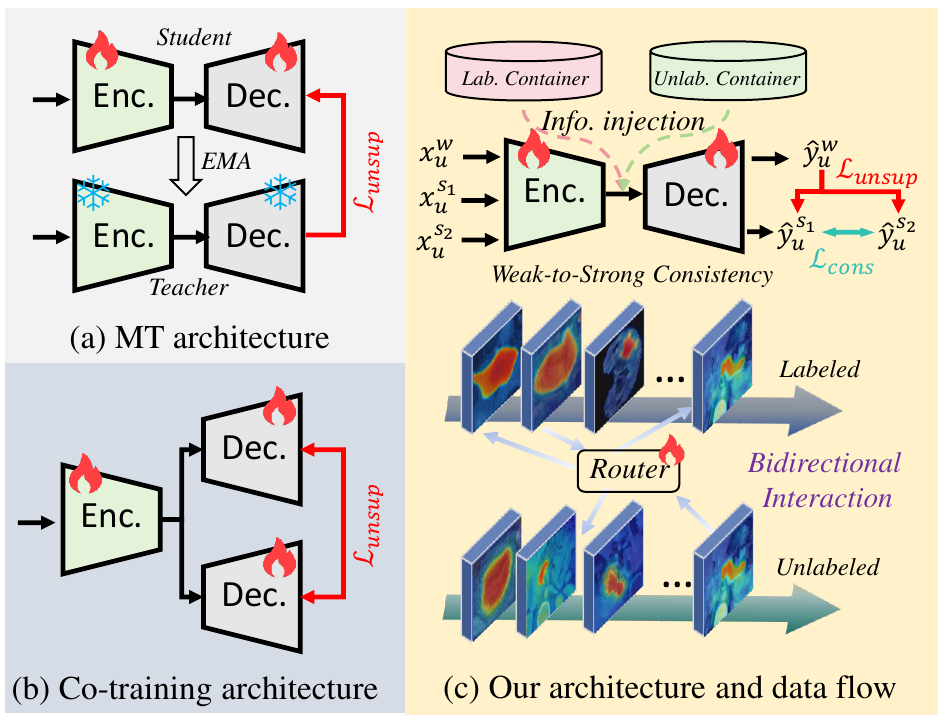} 
\caption{
A comparison of semi-supervised learning paradigms: (a) Mean Teacher framework with student and teacher networks, (b) Co-training framework with two subnets, and (c) Our proposed framework with bidirectional data-stream interaction and semantic-spatial perturbation under weak-to-strong consistency. 
}
\label{fig:fig_1}
\end{figure}

Consistency regularization is a commonly used framework in semi-supervised learning.
For instance, as shown in Fig.~\ref{fig:fig_1}(a), the Mean Teacher (MT)~\cite{tarvainen2017mean} method involves a student network to update parameters using gradient propagation, and a teacher network to update parameters using exponential moving average (EMA). However, in the MT framework, the teacher model is prone to being affected by the accumulated prediction errors of the student model, especially when processing unlabeled data. 
Another widely used strategy is mutual consistency regularization~\cite{wu2021semi}, which employs multi-decoder or dual-branch architectures (Fig.~\ref{fig:fig_1}(b)) to improve model generalization. This is achieved by enforcing consistent predictions across different network heads for the same input. 
While these methods~\cite{peiris2023uncertainty, wang2023mcf} can enhance model stability, they often converge towards similar decision boundaries when processing highly noisy or structurally complex medical images. Moreover, excessive reliance on multiple decoders increases computational complexity and lacks sufficient feature diversity in inter-network knowledge transfer, limiting its effectiveness in mitigating overfitting to specific anatomical structures.

In addition to the challenges mentioned, current semi-supervised methods often train labeled and unlabeled data separately, neglecting their potential interaction. As AllSpark~\cite{wang2024allspark} highlighted that this separation leads to the dominance of labeled data, resulting in low-quality pseudo labels. To address this, this method introduces a channel-wise cross-attention mechanism that regenerates labeled features from unlabeled data. Similarly, SKCDF~\cite{zhang2025semantic} proposes decoupling data streams by separating encoder and decoder roles, enhancing semantic representation learning while reducing the adverse effects of unlabeled data on segmentation.  
However, existing approaches lack bidirectional data interaction and neglect the dual role of feature-level interaction, which acts as both an enhancement and a perturbation between labeled and unlabeled data. Crucially, not all feature channels benefit from ``Reborn" process. Excessive manipulation of channels may introduce redundant information, reducing the model’s representational capacity. 
Without distinguishing each channel's contribution, excessive intervention can introduce noise, compromising model stability and accuracy. Moreover, cross-stream feature interaction remains largely unexplored in 3D medical image segmentation.

To this end, we propose a novel Bidirectional Channel-selective Semantic Interaction (BCSI) framework for semi-supervised medical segmentation. As shown in Fig.~\ref{fig:fig_1}(c), our method employs a paradigm based on weak-to-strong consistency learning, which not only avoids the issues of error accumulation and model structural complexity but also allows for effective interaction between labeled and unlabeled data using a single model structure. Specifically, we employ color jitter and copy-paste as two strong augmentation strategies, perturbing the data in the semantic and spatial domains, respectively. To mitigate the excessive noise and interference from irrelevant features during feature interaction between labeled and unlabeled data, we present the Channel-selective Router (CR) and the Bidirectional Channel-wise Interaction (BCI) strategy. The router dynamically selects which feature channels should interact by learning the importance of each channel, ensuring that only relevant features are activated. The BCI strategy facilitates bidirectional interaction across data streams for the selected channels. Moreover, we construct a feature queue for labeled and unlabeled data, enhancing the model’s long-term memory ability during the training process. Overall, our contributions are summarized as follows:
\begin{itemize}
 
\item We propose BCSI, a novel framework for semi-supervised medical image segmentation that facilitates the interaction between labeled and unlabeled data to enhance the ability of feature representations. BCSI employs a spatial-semantic weak-to-strong consistency learning paradigm, mitigating error accumulation while maintaining architectural simplicity.

\item We propose a channel-selective Router that dynamically identifies critical feature channels during cross-stream interaction between labeled and unlabeled data, mitigating noise interference.

\item We present a bidirectional channel-wise interaction mechanism, primarily focusing on bidirectional interaction between the selected channels and the features stored in the containers.
 
\item Extensive experimental results on multiple benchmark datasets demonstrate that our model significantly improves segmentation accuracy compared to existing semi-supervised segmentation methods.

\end{itemize}


\begin{figure*}[t]
\centering
\includegraphics[width=0.98\textwidth]{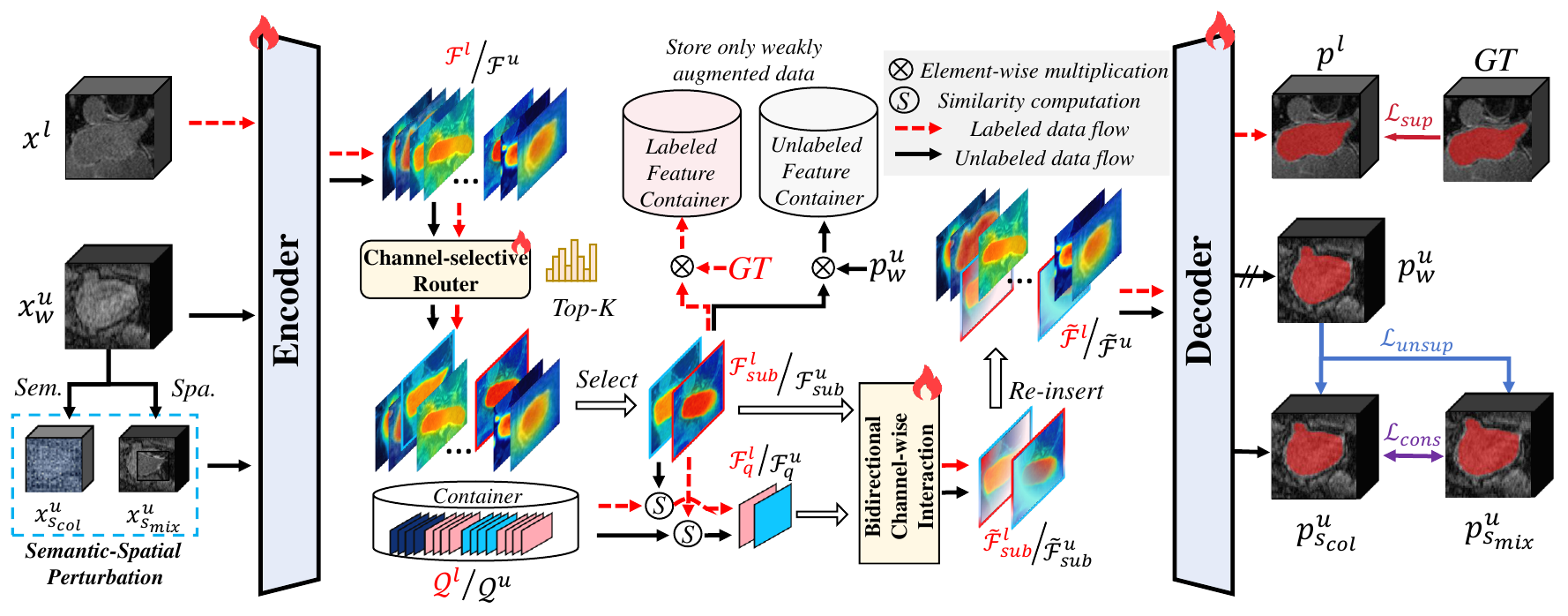}
\caption{
An overview of the proposed BCSI framework, involving an encoder-decoder structure. A channel-selection router then processes the extracted features from labeled and unlabeled data, followed by bidirectional channel-wise interaction for feature enhancement. Finally, the refined features are passed to the decoder under a weak-to-strong consistency learning paradigm. 
}
\label{fig:model}
\end{figure*}

\section{Related Work}

\subsection{Semi-Supervised Medical Image Segmentation }

Semi-supervised medical image segmentation trains with limited labeled and abundant unlabeled data, primarily through pseudo-labeling and consistency learning. 
Pseudo-labeling methods aim to generate pseudo-labels by utilizing the model's predictions on unlabeled data~\cite{han2022effective,seibold2022reference, huang2025learnable, zeng2023ss}.
This work~\cite{han2022effective} proposes to generate pseudo labels via class-wise feature distances and improving segmentation with prior-guided patch sampling.
CoraNet~\cite{shi2021inconsistency} estimates uncertainty via segmentation inconsistency under varying misclassification costs and uses a separate self-training strategy to enhance performance. Consistency learning is widely adopted by enforcing prediction invariance under perturbations and pushing decision boundaries toward low-density regions~\cite{huang2025uncertainty}.
SCO-SSL~\cite{xu2021shadow} enhances prostate segmentation by introducing shadow augmentation to simulate shadow artifacts and shadow dropout to guide boundary inference from shadow-free regions. Tri-U-MT~\cite{wang2021tripled} enhances segmentation by incorporating auxiliary tasks for semantic and shape guidance and introduces a tripled-uncertainty framework for reliable pseudo label generation.
Besides, several studies~\cite{li2023segment, zhang2025semisam+, huang2025text} have been developed to improve the quality of pseudo labels based on foundation models~(\eg~SAM~\cite{kirillov2023segment} and CLIP~\cite{radford2021learning}).

\subsection{Weak-to-Strong Consistency Learning}

Weak-to-strong consistency learning promotes segmentation invariance by applying common perturbations such as Gaussian noise, blur, rotation, scaling, and contrast changes to unlabeled data~\cite{sohn2020fixmatch, kim2022conmatch}. 
These perturbations can be applied at both the input~\cite{huang2022semi} and feature~\cite{xu2022learning} levels. 
UniMatch~\cite{yang2023revisiting} enhances consistency training by introducing an auxiliary feature perturbation stream and a dual-stream image-level perturbation strategy, enabling stronger and more diverse supervision signals.
\cite{bortsova2019semi} proposed enforcing segmentation consistency under elastic deformations using a Siamese network with shared weights and a composite loss that combines supervised and consistency terms.
UDC-Net~\cite{li2021dual} introduces a dual-consistency learning scheme combining image transformation and feature perturbation consistency, guided by quantified uncertainty to improve lesion segmentation.

\section{Methodology}

\textbf{Overview}. In the semi-supervised setting, we have a labeled dataset $\mathcal{D}^L=\{(x^l_i, y^l_i)\}_{i=1}^{N_l}$ and an unlabeled dataset $\mathcal{D}^U=\{x^u_i\}_{i=N_l+1}^{N_l+N_u}$, where $N_l$ and $N_u$ denote the numbers of labeled and unlabeled samples, respectively, with $N_l \ll N_u$. $x_i \in \mathbb{R}^{H \times W \times D}$ represents the input volume, and $y_i \in \{0, 1\}^{H \times W \times D}$ denotes the corresponding ground-truth map. 
Fig.~\ref{fig:model} shows the architecture of our framework. Specifically, we first apply two strong augmentations and one weak augmentation to the input data, then pass them through the encoder to extract features. After that, the router selects the top-$k$ channels for cross-data stream interaction. Finally, the features are reinserted into the original channels and stored in their containers.

\subsection{Semantic-Spatial Perturbation}

In semi-supervised learning, data augmentation plays a crucial role in enhancing model generalization~\cite{yang2023revisiting}. Departing from conventional approaches like multi-decoder or dual-stream architectures, we propose a Semantic-Spatial Perturbation (SSP) method to bolster model robustness. For semantic perturbation, we leverage color jitter~\cite{cubuk2020randaugment}, which introduces random variations in brightness, contrast, and Gaussian noise. Formally, this can be expressed as follows:
\begin{equation} 
x^u_{s_{col}} = \alpha \cdot x^u + \beta + \mathcal{N}(\mu, \sigma^2),
\end{equation}
where $x^u_{s_{col}}$ represents the color jitter augmented unlabeled data.
$\alpha$ and $\beta$ denote the randomly sampled contrast factor and brightness offset, respectively. $\mathcal{N}(\mu, \sigma^2)$ represents Gaussian noise with mean $\mu$ and standard deviation $\sigma$.
For spatial perturbations, we adopt a copy-paste~\cite{bai2023bidirectional} strategy by randomly initializing a binary mask $\mathcal{M}$ to spatially blend volumes from both labeled and unlabeled data. This operation introduces structural diversity by disrupting spatial continuity and generating composite inputs, which can be formulated by
\begin{equation}
{x}^{u}_{s_{mix}} = \mathcal{M} \odot x^{l} + (1 - \mathcal{M}) \odot x^{u},
\end{equation}
where ${x}^{u}_{s_{mix}}$ denotes the copy-paste augmented unlabeled data.
$\odot$ denotes element-wise multiplication. 
Then, the strongly augmented images and weakly augmented images ${x}^{u}_w$ are fed into the model, which can be expressed as:
\begin{equation}
p_{s_{col}}^u = {f}({x}^{u}_{s_{col}};\theta_f), p_{s_{mix}}^u = {f}({x}^{u}_{s_{mix}};\theta_f),  p_{w}^u = {f}({x}^{u}_w;\theta_f),
\end{equation}
where $p_{s_{col}}^u$ and $p_{s_{mix}}^u$ represent the predictions for the color-jitting and copy-paste data augmentations, respectively. $\theta_f$ represents the parameters of $f(\cdot)$. $p_{w}^u$ is the prediction for the weakly augmented unsupervised data. All the data share the same model parameters.
We use $p_{w}^u$ as pseudo-labels for the unsupervised loss and apply consistency loss between the two strongly augmented predictions $p_{s_{col}}^u$ and $p_{s_{mix}}^u$.
The detailed formulation of the losses in this part will be discussed collectively in Sec. ``Loss Function".

\subsection{Channel-selective  Router}

To mitigate the insufficient feature-level interaction between labeled and unlabeled data while narrowing their distribution gap, we propose the Channel-selective Router (CR). Our key insight is that feature exchange not only enhances representations but also acts as a structured perturbation to improve model robustness.
Unlike prior works that indiscriminately mix features across all channels, CR employs a lightweight and learnable routing mechanism to dynamically select and perturb the most informative channels.

Specifically, given the input volumes from the labeled and unlabeled sets, we first obtain the corresponding features $\mathcal F^l \in \mathbb{R}^{C \times h \times w \times d}$ and $\mathcal F^u \in \mathbb{R}^{C \times h \times w \times d}$  through a shared encoder, where $C$, $h$, $w$, and $d$ represent the number of channels, height, width, and depth, respectively. These features are then fed into a lightweight router $\mathcal{G}(\cdot)$ to generate channel-wise activation scores, denoted as $\textbf{s} \in \mathbb{R}^C$, which quantify the importance or perturbation sensitivity of each channel. The process can be described as follows:
\begin{equation}
\textbf{s} = \mathcal{G}(\{\mathcal F^l, \mathcal F^u\}; \theta_{\mathcal{G}}),
\end{equation}
where $\theta_{\mathcal{G}}$ is the learnable parameter of the router.
Subsequently, we construct a sparse channel mask $\mathcal{R} \in \mathbb{R}^C$ to indicate the top‑$K$ most activated channels, which can be expressed by
\begin{equation}
\mathcal{R} = \delta \left( \mathbf{s} \geq \tau_K(\mathbf{s}) \right),
\end{equation}
where $\delta(\cdot)$ denotes the channel-level threshold function, and $\tau_K(\cdot)$ is the threshold corresponding to the $K$-largest value in $\mathbf{s}$. 
Then, we select the features to be perturbed using the sparse channel mask $\mathcal{R}$, which can be expressed as follows:
\begin{equation}
\mathcal{F}^l_{sub} = \mathcal{F}^l  \odot  \mathcal{R}^l, \mathcal{F}^u_{sub} = \mathcal{F}^u  \odot  \mathcal{R}^u, 
\end{equation}
where ${\mathcal{F}}^l_{sub} \in \mathbb{R}^{K \times h \times w \times d}$ and ${\mathcal{F}}^u_{sub} \in \mathbb{R}^{K \times h \times w \times d}$ respectively represent the features selected from labeled and unlabeled data. 
$\mathcal{R}^l$ and $\mathcal{R}^u$ represent the corresponding sparse channel masks, respectively. Note that ${\mathcal{F}}^l_{sub} \subset \mathcal{F}^l$ and ${\mathcal{F}}^u_{sub} \subset \mathcal{F}^u$ are subsets of $\mathcal{F}^l$ and $\mathcal{F}^u$, respectively, obtained through the masks $\mathcal{R}^l$ and $\mathcal{R}^u$. Finally, we perform channel-wise interaction on the selected features.

\subsection{Bidirectional Channel-wise Interaction}
To explore the semantic relationships between different data streams, we perform bidirectional feature interaction between labeled and unlabeled data, reactivating features to achieve deep information interaction.
Firstly, two feature containers $\mathcal{Q}^l \in \mathbb{R}^{M \times L}$ and $\mathcal{Q}^u \in \mathbb{R}^{M \times L}$ are randomly initialized to store the selected features from the labeled and unlabeled data, where $M$ and $L$ denote the maximum length of the containers and the length of a single channel, respectively.
Then, we compute the similarity between the selected channels and the features stored in the containers, from which the most similar features to each selected channel are retrieved. This process can be expressed as follows:
\begin{equation}
\scalebox{0.95}{$
\left\{
\begin{aligned}
& \mathcal{F}^l_q = \{\arg \max_{f_q \in \mathcal{Q}^l} \text{Sim}(\mathcal{F}^l_{sub, k}, f_q)\}_{k=1}^K, \\
& \mathcal{F}^u_q = \{\arg \max_{f_q \in \mathcal{Q}^u} \text{Sim}(\mathcal{F}^u_{sub, k}, f_q)\}_{k=1}^K, \\
\end{aligned}
\right.
$}
\end{equation}
where $\text{Sim}(\cdot,\cdot)$ represents the cosine similarity function. $\mathcal{F}^l_{sub, k}$ and $\mathcal{F}^u_{sub, k}$ represent the $k$-th channel feature of $\mathcal{F}^l_{sub}$ and $\mathcal{F}^u_{sub}$, respectively.
$\mathcal{F}^l_q  \in \mathbb{R}^{K \times h \times w \times d} $ and $\mathcal{F}^u_q  \in \mathbb{R}^{K \times h \times w \times d}$ represent the most similar features selected from the containers, respectively.
Subsequently, we perform mutual feature perturbation between the labeled and unlabeled data flow, which can be expressed as follows:
\begin{equation}
\scalebox{0.95}{$
\left\{
\begin{aligned}
& \tilde{\mathcal{F}}_{sub}^l = \sigma\left(\mathbf{Q}(\mathcal{F}_{sub}^l) \cdot \mathbf{K}({\mathcal{F}}_q^u)^\top / \sqrt{d}\right) \cdot \mathbf{V}(\mathcal{F}_q^u) + \mathcal{F}_{sub}^l, \\
& \tilde{\mathcal{F}}_{sub}^u = \sigma\left(\mathbf{Q}(\mathcal{F}_{sub}^u) \cdot \mathbf{K}({\mathcal{F}}_q^l)^\top / \sqrt{d}\right) \cdot \mathbf{V}(\mathcal{F}_q^l) + \mathcal{F}_{sub}^u,
\end{aligned}
\right.
$}
\label{eq:8}
\end{equation}
where $\mathbf{Q}$, $\mathbf{K}$, and $\mathbf{V}$ represent the linear projections. $\sqrt{d}$ is a scaling factor, and $\sigma(\cdot)$ represents the softmax activation function. Finally, the perturbed features are re-inserted into the original features, guided by the sparse channel mask $\mathcal{R}$, which can be expressed as follows:
\begin{equation}
\scalebox{0.95}{$
\left\{
\begin{aligned}
& \tilde {\mathcal{F}^{l}} = \tilde{\mathcal{F}}_{sub}^l  \odot  \mathcal{R}^l + \mathcal{F}^{l} \odot (1- \mathcal{R}^l), \\
& \tilde {\mathcal{F}^{u}} = \tilde{\mathcal{F}}_{sub}^u  \odot  \mathcal{R}^u + \mathcal{F}^{u} \odot (1- \mathcal{R}^u),
\end{aligned}
\right.
$}
\end{equation}
where $\tilde {\mathcal{F}}^{l}  \in \mathbb{R}^{C \times h \times w \times d}$ and $\tilde {\mathcal{F}}^{u}  \in \mathbb{R}^{C \times h \times w \times d}$ represent the features of the labeled and unlabeled data after interaction, respectively.
This bidirectional channel-wise interaction strategy enables targeted interaction between the feature distributions of labeled and unlabeled data, with the router learning to modulate only the most relevant semantic features, avoiding unnecessary interference. 
It is worth noting that $\mathcal{Q}^l$ and $\mathcal{Q}^u$ are feature queues operating on a first-in-first-out basis. When full, they discard the oldest entries. At each iteration, features from labeled data are multiplied by ground-truth class labels while unlabeled data features are multiplied by pseudo-labels, and the obtained features are appended to the respective queues.

\subsection{Loss Function}
\label{sec:loss}

The total loss function includes supervised loss, unsupervised loss, and consistency loss, which can be expressed by
\begin{equation}
\mathcal{L}_{total} = \mathcal{L}_{sup} + \mathcal{L}_{cons} + \lambda_u \mathcal{L}_{unsup}
\end{equation}
where $\lambda_u$ employs a Gaussian warm-up function $\lambda_{u}(t) = \beta * e^{-5(1-t/t_{max})^2}$ with $\beta = 0.1$, and $t_{max}$ denotes the maximum number of iterations. 
To encourage the model to focus more attention on challenging segmentation regions, such as boundaries between different classes, we introduce uncertainty values as weighting guidance~\cite{zhao2024textpolyp} in the segmentation loss calculation. 
In the following losses, we apply the weighted segmentation loss $\mathcal{L}_{seg}$ throughout, which can be expressed as follows:
\begin{equation}
\mathcal{L}_{seg}(p, y, \mathcal{W}) = \mathcal{L}_{BCE} + \mathcal{L}_{IoU},
\end{equation}
where $p$, $y$, and $\mathcal{W}$ represent the prediction, ground-truth, and uncertainty weight map, respectively.

\begin{table*}[!t]
  \centering
  \scriptsize
  \renewcommand{\arraystretch}{1.0}
  \setlength\tabcolsep{1.4pt}

\begin{tabular}{c|c|cccc|cccc|c|cccc|cccc}
    \toprule
\multicolumn{1}{c|}{\multirow{2}{*}{Method}} &  & \multicolumn{4}{c|}{10\% / 8 labeled data} & \multicolumn{4}{c|}{20\% / 16 labeled data} &  & \multicolumn{4}{c|}{10\% / 25 labeled data} & \multicolumn{4}{c}{20\% / 50 labeled data} \\
\cline{3-10}\cline{12-19}
   &  & Dice~$\uparrow$ & Jaccard~$\uparrow$ & 95HD~$\downarrow$ & ASD~$\downarrow$    & Dice~$\uparrow$ & Jaccard~$\uparrow$ & 95HD~$\downarrow$ & ASD~$\downarrow$ & & Dice~$\uparrow$ & Jaccard~$\uparrow$ & 95HD~$\downarrow$ & ASD~$\downarrow$    & Dice~$\uparrow$ & Jaccard~$\uparrow$ & 95HD~$\downarrow$ & ASD~$\downarrow$    \\
\midrule                     
{VNet~(SupOnly)}& \multirow{12}{*}{\begin{sideways}Left Atrium\end{sideways}}  &   82.74 & 71.72  &  13.35 & 3.26 &  84.93 & 75.87  & 14.50 & 4.30 & \multirow{12}{*}{\begin{sideways}BraTS-2019\end{sideways}} &  74.43  &  61.86  &   37.11   &  2.79      &  80.16        &    71.55 &   22.68 &   3.43 \\
\cline{3-10}\cline{12-19}
UAMT~\cite{yu2019uncertainty}  &   & 87.79 & 78.39 & 8.68 & 2.12   & 88.97 & 80.38 & 8.18 & 1.92 & &  79.49  &   69.22  &  11.93  &  1.93 & 79.72  & 69.46 & 11.26  & 1.97   \\
DTC~\cite{luo2021semi} &   & 87.51 & 78.17 & 8.23 & 2.36 & 88.32 & 79.34 & 8.72 & 2.02  & & 77.54   &  66.91   & 12.16  & 2.84  &  82.38  &  72.15  &  11.00 & 2.16 \\
MC-Net~\cite{wu2021semi} &    & 87.62 & 78.25 & 10.03 & 1.82 & 90.37 & 82.52 & 6.77 & 1.71  & & 81.52   &  71.89  &  13.26  &  4.56 & 83.79  &   73.69  &  9.65 & 1.76 \\
MC-Net+ \cite{wu2022mutual} &     & 88.89 & 80.15 & 8.01 & 1.90 & 90.31 & 82.41 & 6.61 & 1.62  & &   79.27  &   68.97& 14.89  & 4.91  & 83.47 &  72.89  &  9.69  & 1.92\\
URPC \cite{luo2022semi} &  & 86.92 & 77.03 & 11.13 & 2.28 & 87.68 & 78.36 & 9.39 & 3.52  &  & 82.59  &  72.11     &  13.88    &  3.72       &  82.93   & 72.57  & 15.93   &  4.19\\
MCF~\cite{wang2023mcf} &  & 87.06 & 77.83 & 7.81 & 2.67 & 88.71 & 80.41 & 6.32 & 1.90  & &    78.83     &   68.49   &   12.25    &  1.92  &  80.07 &  69.55 &  10.65  & 2.00\\
BCP~\cite{bai2023bidirectional} &   & {89.62} & {81.31} & 6.81 & {1.76} & 90.38 & 82.57 & 6.68 & 1.76  & & 78.64    & 68.59   &  12.88  &2.81  & 80.20  &  69.66 &  10.52 &  2.08 \\
BS-Net~\cite{he2023bilateral} & & 82.53  & 71.27   & 12.48  &  3.30 &  90.43 &  - &  6.21 & 1.63  & &  79.93 &  69.37  & 10.88  & 2.05 &    82.03  &   71.87   &  10.26 &  1.96\\

MLRP~\cite{su2024mutual} & &  89.86 &  81.68 &  6.91 & 1.85 & \underline{91.02}  & \underline{83.62}   & 5.78  &  1.66  & & {84.29}  & {74.74}  &  {9.57}  & 2.55 &  \underline{85.47}      & \underline{76.32} &  \underline{7.76} & 2.00\\
SKCDF~\cite{zhang2025semantic} & &  89.42 & 80.46 & 8.18 &  2.38 &  90.65 & 82.98 & 6.70 &  \underline{1.61} & &  83.00 & 73.08 & 10.28 &  \underline{1.78} &  84.40 & 74.78 & 9.31 &  \underline{1.68}\\
UnCo~\cite{zeng2025uncertainty} &   &  \underline{90.37} & \underline{82.54} & \underline{6.11} &  \underline{1.74}  &  90.91 & 83.40 & \underline{5.36} &  \underline{1.61} &  &  \underline{85.09} & \underline{75.64} & \underline{8.63} &  1.89 & 85.16 & 75.04 & 8.41 &  1.74 \\
\textbf{Ours} & &   \textbf{91.07}  &  \textbf{83.67}   &   \textbf{5.57} &  \textbf{1.53}  &    \textbf{91.84}  &  \textbf{84.98}   &   \textbf{5.06} &  \textbf{1.38}  & &   \textbf{86.17}   &   \textbf{77.10}   & \textbf{8.43}   &  \textbf{1.37}  &    \textbf{86.86}  &  \textbf{77.80}   &   \textbf{7.62} &  \textbf{1.26} \\
\bottomrule
\end{tabular}
\caption{
Results on the LA and BraTS-2019 datasets (Best and second-best results are highlighted in \textbf{Bold} and \underline{Underline}).}
\label{tab_Comparison_1}
\end{table*}

We apply two different strong augmentations and one weak augmentation to the unsupervised data, and take the predictions from the weak augmentation as pseudo-labels to supervise the strongly augmented data. The unsupervised loss with the weak-to-strong consistency strategy can be represented as follows:
\begin{equation}
\mathcal{L}_{unsup} = \mathcal{L}_{seg}(p_{s_{col}}^u, p_{w}^u, \mathcal{W}_{s_{col}}^u) + \mathcal{L}_{seg}(p_{s_{mix}}^u, p_{w}^u, \mathcal{W}_{s_{mix}}^u),
\end{equation}
where $\mathcal{W}_{s_{col}}^u$ and $\mathcal{W}_{s_{mix}}^u$ represent their corresponding uncertainty weights. Note that for the strong augmentation operation involving spatial perturbation, we perform spatial restoration before calculating the loss. In addition, we perform cognitive consistency learning on the unlabeled data with different strong augmentations, which can be expressed as follows:
\begin{equation}
\mathcal{L}_{cons} =  \frac{1}{H \times W}\sum_{i\in H ,j\in W} ( p_{s_{col}}^u(i,j) - p_{s_{mix}}^u(i,j))^2.
\end{equation}

Finally, we apply supervised loss to the labeled data. In this process, we also perform two types of strong augmentations and one weak augmentation on the labeled data to enable the model to adapt to this augmentation strategy, which can be expressed as follows:
\begin{equation}
\begin{aligned}
\mathcal{L}_{sup} = & \mathcal{L}_{seg}(p_{s_{col}}^l, y^l, \mathcal{W}_{s_{col}}^l) + \mathcal{L}_{seg}(p_{s_{mix}}^l, y^l, \mathcal{W}_{s_{mix}}^l) \\
& + \mathcal{L}_{seg}(p_{w}^l, y^l, \mathcal{W}_{w}^l),
\end{aligned}
\end{equation}
where $p_{s_{col}}^l$ and $p_{s_{mix}}^l$ represent the predictions for the strongly augmented labeled data, and $\mathcal{W}_{s_{col}}^l$ and $\mathcal{W}_{s_{mix}}^l$ are their corresponding uncertainty weights. $p_{w}^l$ and $\mathcal{W}_{w}^l$ denote the predictions and uncertainty weights for the weakly augmented labeled data, respectively. Additionally, $y^l$ represents the ground truth of the labeled data.

\section{Experiments}

\subsection{Experimental Setup}

\textbf{Datasets}. We conduct comparison experiments on three widely used datasets. 
$\bullet$ The Left Atrium~(LA) dataset~\cite{xiong2021global} is the benchmark for the 2018 atrial segmentation challenge. It includes 100 3D gadolinium-enhanced magnetic resonance image volumes, all with corresponding labels. Following the same experimental setup~\cite{bai2023bidirectional}, we divide the dataset into 80 samples for training and 20 samples for testing in our experiments.
$\bullet$ The Pancreas-CT dataset~\cite{roth2015deeporgan} consists of 82 3D abdominal CT scans with a resolution of $512 \times 512$ pixels, and slice thicknesses ranging from 1.5 mm to 2.5 mm. Following the previous setting~\cite{luo2021semi}, we partition the dataset into 62 training samples and 20 testing samples.
$\bullet$ The BraTS-2019 dataset~\cite{hdtd-5j88-20} contains multi-modal MRI scans from 335 glioma patients, with four imaging sequences: T1, T1CE, T2, and FLAIR. Following the configuration in~\cite{ssl4mis2020}, we use only FLAIR images for tumor segmentation. The dataset is split into 250 training, 25 validation, and 60 testing samples.

\textbf{Implementation Details.}
All experiments are performed in a PyTorch 1.8.1 environment with CUDA 11.2, utilizing an NVIDIA 4090 GPU. To maintain consistency, we retain the same backbone selection as in previous work, opting for VNet~\cite{milletari2016v}. 
We utilize the SGD optimizer with an initial learning rate of $0.01$, a momentum of $0.9$, and a weight decay of $0.0005$.
The batch size is set to $4$.
The total number of iterations is set to $30k$.
During the feature-level interaction process, we select the top $64$ channels.
The maximum length $k$ of the container is set to 2560.
Based on previous settings, during the training process, the cropping size for the LA dataset is $112 \times 112 \times 80$, while for the other datasets, the cropping size is $96 \times 96 \times 96$. 
During the testing phase, no perturbations or augmentation operations are applied.

\textbf{Evaluation Metrics.}
All of our comparative experiments utilize four evaluation metrics, namely Dice coefficient, Hausdorff distance (95HD), Intersection over Union (IoU), and Average Surface Distance (ASD). 

\subsection{Comparison With State-of-the-art Methods}

\textbf{Comparison Methods.} The proposed method is evaluated against 11 state-of-the-art semi-supervised medical image segmentation methods, including UAMT~\cite{yu2019uncertainty}, DTC~\cite{luo2021semi}, MC-Net~\cite{wu2021semi}, MC-Net+~\cite{wu2022mutual}, URPC~\cite{luo2022semi}, MCF~\cite{wang2023mcf}, BCP~\cite{bai2023bidirectional}, BS-Net~\cite{he2023bilateral}, MLRP~\cite{su2024mutual}, SKCDF~\cite{zhang2025semantic}, and UnCo~\cite{zeng2025uncertainty}.

\begin{figure*}[t]
\centering
\includegraphics[width=0.999\textwidth]{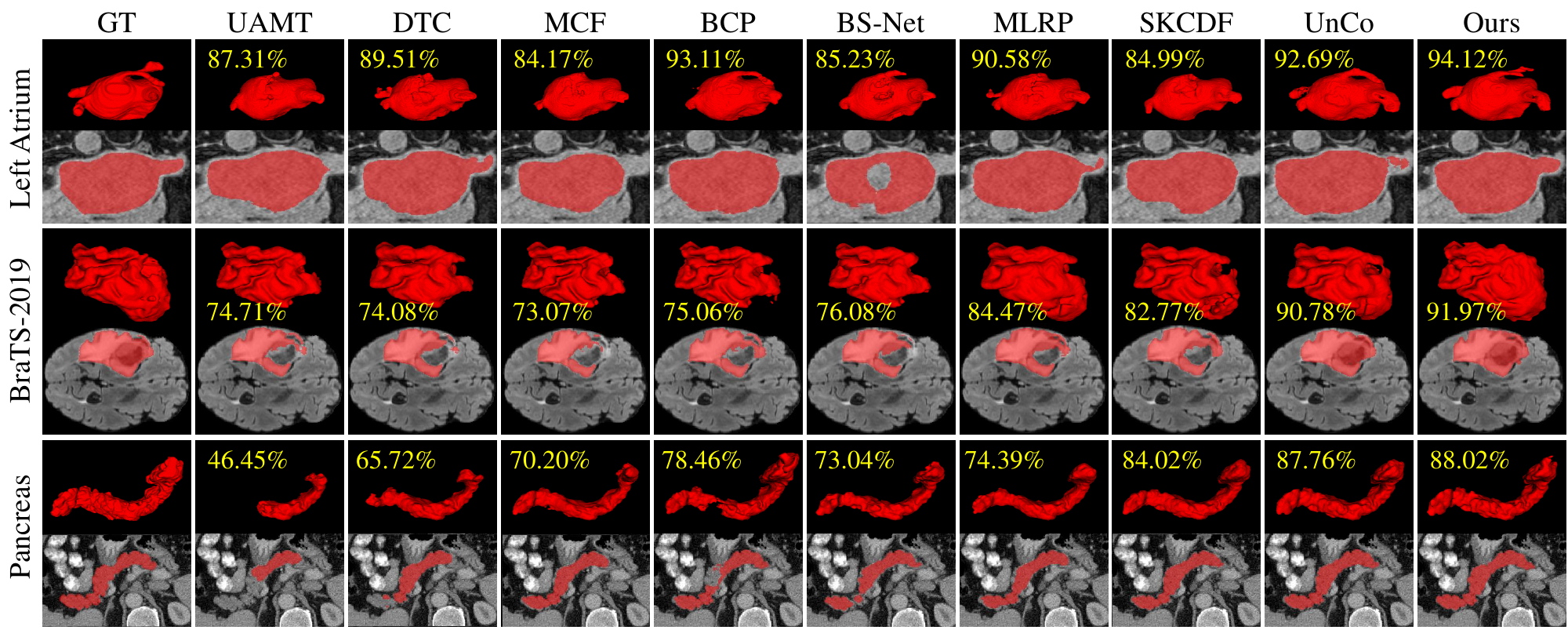} 
\caption{
Visualization results of our method compared to other semi-supervised methods on the LA, BraTS-2019, and Pancreas datasets. The yellow numbers represent the Dice score of the currently displayed sample.
}
\label{fig:vis_res}
\end{figure*}

\textbf{Quantitative and Qualitative Comparisons.}
Table \ref{tab_Comparison_1} presents the quantitative results of our method on the LA and BraTS-2019 datasets, where our semi-supervised approach achieves satisfactory performance.
On the BraTS-2019 dataset with a $10\%$ labeled ratio, our method improves the Dice score from $85.09\%$ to $86.17\%$ compared to the second-best model (UnCo), and reduces the 95HD from $1.89$ to $1.37$. Additionally, when compared to the SKCDF network, which also uses a dual-stream interaction structure, the Dice score increases by $3.17\%$. Furthermore, it continues to achieve the best results at other labeled ratios.
Table~\ref{tab_Comparison_pancreas} shows the comparison of our method with other semi-supervised approaches on the Pancreas dataset. In the case of $10\%$ labeled data, compared to UnCo, the Dice score increases from $78.53\%$ to $80.41\%$, and the 95HD decreases from $7.36$ to $6.33$. Fig.~\ref{fig:vis_res} shows the qualitative results of our method and other comparison methods. It can be observed that our method can still accurately capture the target regions, particularly in handling complex structures and blurred boundaries.

\begin{table}[t!]
  \centering
  \scriptsize
  \renewcommand{\arraystretch}{1.0}
  \setlength\tabcolsep{1.8pt}
  \begin{tabular}{c|cccc|cccc}
\toprule
\multirow{2}{*}{Method}  & \multicolumn{4}{c|}{10\% / 6 labeled data} & \multicolumn{4}{c}{20\% / 12 labeled data} \\
\cline{2-9}
  & Dice~$\uparrow$ & Jaccard~$\uparrow$ & 95HD~$\downarrow$ & ASD~$\downarrow$ & Dice~$\uparrow$ & Jaccard~$\uparrow$ & 95HD~$\downarrow$ & ASD~$\downarrow$  \\
 \midrule
VNet  &   54.94& 40.87  &  47.48 & 17.43   & 75.07  &   61.96 & 10.79 & 3.31      \\
\cline{2-9}
UAMT &  66.44 &  52.02   &  17.04       &   3.03  &   76.10   & 62.62      &    10.87    &   2.43 \\
DTC   & 66.58   &  51.79   &   15.46  & 4.16   &  76.27   &   62.82  & 8.70 & 2.20    \\
MC-Net  & 69.07 & 54.36 &    14.53     &   2.28   &   78.17         &     65.22 &    6.90     &  1.55    \\
MC-Net+ & 70.00 &  55.66  & 16.03  & 3.87  & 79.37  &  66.83    & 8.52 & 1.72    \\
URPC  &  73.53 &   59.44 &  22.57  & 7.85 &    80.02   & 67.30  &   8.51  &  1.98     \\
MCF  & 67.71 &  53.83 &  17.17  &  {2.34} &  75.00 &  61.27 &  11.59  &  3.27    \\
BCP & 75.17   &  60.99 &  10.49 & 2.32  &  {82.91} &  {70.97} & {6.43} & 2.25    \\
BS-Net &  64.61   &  50.02  &  24.74  &  5.28  & 78.93   & 65.75       &8.49  &  2.20      \\
MLRP  & {75.93} &  {62.12}   &  {9.07}   &  {1.54}   & {81.53} & 69.35  &  6.81 & \underline{1.33}    \\
SKCDF & 70.71  & 56.48  & 19.84  & 1.37  &  78.30 & 65.27 & 9.34 &  1.38 \\
UnCo & \underline{78.53}  & \underline{65.20}  & \underline{7.36}  & \underline{1.33}  &  \underline{81.93} & \underline{69.77} & \underline{5.34} &  1.36 \\
\textbf{Ours} & \textbf{80.41}  & \textbf{67.70} & \textbf{6.33}  & \textbf{1.32}   & \textbf{83.24}  & \textbf{71.61} & \textbf{4.87} & \textbf{1.16}  \\
\bottomrule
\end{tabular}
\caption{Quantitative results on the Pancreas-CT dataset.}
\label{tab_Comparison_pancreas}
\end{table}

\subsection{Ablation Studies}

\begin{table}[t!]
  \centering
  \scriptsize
  \renewcommand{\arraystretch}{1.0}
  \setlength\tabcolsep{1.8pt}
\begin{tabular}{c|ccc|cccc|cccc}
  \toprule
\multirow{2}{*}{Baseline} & \multicolumn{3}{c|}{BCSI} & \multicolumn{4}{c|}{Left Atrium~(20\%)} & \multicolumn{4}{c}{BraTS-2019~(20\%)} \\
\cline{2-12}
& SSP    & BCI  & CR   & Dice & Jaccard & 95HD & ASD   & Dice & Jaccard & 95HD & ASD  \\
\midrule
\checkmark &     &    &       &    88.60    &  79.82     &  10.98      &  2.80     &    82.86    & 73.24       &    10.31    &  1.72    \\
\checkmark &   \checkmark   &    &    &   90.58     &   82.96    &   6.43     &  1.58     &  85.59     &   76.34   &   8.51 & 1.42     \\
\checkmark &     &   \checkmark   &   &   89.23     &   80.79    &   7.50     &   2.26    &  84.12      & 74.61      &    9.65    &  1.62    \\
\checkmark &     & \checkmark  &  \checkmark      &  90.22      &   82.40    &    6.49    &  1.85     &  84.72      &   75.44    &   9.04     &  1.32    \\
\checkmark & \checkmark    & \checkmark  &  \checkmark      &    \textbf{91.84}  &  \textbf{84.98}   &   \textbf{5.06} &  \textbf{1.38}      &    \textbf{86.86}  &  \textbf{77.80}   &   \textbf{7.62} &  \textbf{1.26}     \\
\bottomrule
\end{tabular}
  \caption{Ablation study on key components in our method.}
    \label{tab:tab_AS_Key}
\end{table}

We conduct ablative studies on the key components and all proposed strategies within the BCSI framework. All experiments are performed on the LA and Pancreas datasets with a $20\%$ labeled ratio.

\textbf{Effectiveness of Key Components.}
Table~\ref{tab:tab_AS_Key} shows the results of our quantitative analysis to explore the effectiveness of the key components in our method. Notably, without the SSP strategy, the weak-to-strong consistency learning framework cannot be applied, and we substitute it with the MT structure. Additionally, applying only the BCI module without the CR module implements a full channel interaction strategy. As shown in Table~\ref{tab:tab_AS_Key}, the SSP structure based on weak-to-strong consistency outperforms traditional semi-supervised structures like MT. Furthermore, applying the BCI module within the MT structure improves performance, validating the effectiveness of the dual-stream interaction strategy. Building on this, the introduction of channel routing further improves performance. On the LA dataset with $20\% $labeled data, the Dice score increases from $89.23\%$ to $90.22\%$, and when all strategies are applied, the performance rises to $91.84\%$.

\textbf{Effectiveness of Different Settings in SSP}. We employ an ablation study to quantify the impact of different augmentation settings in our SSP strategy, and Table~\ref{tab:tab_AS_augmentation} presents ablative results. Notably, when only weak augmentation is applied, the MT framework is used as a substitute with keeping all other strategies. As shown in Table~\ref{tab:tab_AS_augmentation}, it can be observed that our SSP with one weak and two strong augmentation settings achieves the best performance. This indicates our strategy fully leverages changes in the semantic and spatial domains to enhance model performance.

\begin{table}[t!]
  \centering
  \scriptsize
  \renewcommand{\arraystretch}{1.0}
  \setlength
  \tabcolsep{2.5pt}
\begin{tabular}{ccc|cccc|cccc}
\toprule
\multicolumn{3}{c|}{Settings} & \multicolumn{4}{c|}{Left Atrium~(20\%)} & \multicolumn{4}{c}{BraTS-2019~(20\%)} \\
\hline
Weak        & Col.       & Mix.   & Dice  & Jaccard  & IoU  & ASD    & Dice & Jaccard & 95HD  & ASD     \\
\midrule
   \checkmark &        &        &  90.22      &   82.40    &    6.49    &  1.85     &  84.72      &   75.44    &   9.04     &  1.32      \\
  \checkmark &   \checkmark       &         &  90.87    &  83.33   &   6.38  &  1.70   &    85.80  &  76.69    &   9.18  & 1.49    \\
\checkmark  &         &  \checkmark   &   90.97   &   83.50  &  6.49   &  1.50   &   86.16   &  77.30    &  8.44   &   1.27  \\
\checkmark &   \checkmark      &  \checkmark      &    \textbf{91.84}  &  \textbf{84.98}   &   \textbf{5.06} &  \textbf{1.38}     &    \textbf{86.86}  &  \textbf{77.80}   &   \textbf{7.62} &  \textbf{1.26}     \\
\bottomrule
\end{tabular}
\caption{Ablation study of different settings in the SSP.}
\label{tab:tab_AS_augmentation}
\end{table}

\begin{table}[h]
  \centering
  \scriptsize
  \renewcommand{\arraystretch}{1.0}
  \setlength\tabcolsep{3.0pt}
\begin{tabular}{c|cccc|cccc}
\hline
\multirow{2}{*}{Settings} & \multicolumn{4}{c|}{Left Atrium~(20\%)} & \multicolumn{4}{c}{BraTS-2019~(20\%)} \\
\cline{2-9}
& Dice & Jaccard & 95HD & ASD   & Dice & Jaccard & 95HD & ASD  \\
\hline
Lab. $\times$  Unlab.       &   90.58     &   82.96    &   6.43     &  1.58     &  85.59     &   76.34   &   8.51 & 1.42     \\
Lab. $\rightarrow$  Unlab.       &    91.42    &   84.29    &  5.34      &   1.50    &  86.13  & 77.04  &   8.17   &   1.34   \\
Lab. $\leftarrow$  Unlab.        &    91.39    & 84.24      &   5.60     &  1.55     &   86.05     & 76.72      &   8.30     &  1.48    \\
Lab. $\leftrightarrow$  Unlab.   &    \textbf{91.84}  &  \textbf{84.98}   &   \textbf{5.06} &  \textbf{1.38}      &    \textbf{86.86}  &  \textbf{77.80}   &   \textbf{7.62} &  \textbf{1.26}   \\
\hline
\end{tabular}
\caption{Ablation study of interactive data flow direction.}
\label{tab:tab_AS_dataflow}
\end{table}

\textbf{Effectiveness of Bidirectional Interaction}. We investigate the effects of different interaction directions between the labeled and unlabeled data flows. Notably, ``Lab. $\rightarrow$  Unlab." refers to the process of using labeled data to augment unlabeled data, which corresponds to executing only the second half of Eq.~(\ref{eq:8}), with the reverse applying similarly. 
As shown in Table~\ref{tab:tab_AS_dataflow}, perturbations from labeled data to unlabeled data perform better than those from unlabeled data to labeled data. This is because the features of labeled data are generally more stable, providing more reliable information to guide the learning process of unlabeled data. More importantly, bidirectional interaction achieves the best performance, by exploiting the potential relationship between labeled and unlabeled data to enhance their complementarity.

\begin{table}[t!]
  \centering
  \scriptsize
  \renewcommand{\arraystretch}{1.0}
  \setlength\tabcolsep{3.5pt}

\begin{tabular}{c|cccc|cccc}
\toprule
\multirow{2}{*}{Settings} & \multicolumn{4}{c|}{Left Atrium~(20\%)} & \multicolumn{4}{c}{BraTS-2019~(20\%)} \\
\cline{2-9}
                          & Dice & Jaccard & 95HD & ASD   & Dice & Jaccard & 95HD & ASD  \\
\midrule 
$K=32$     &     91.66   &   84.68    &   5.88     &  1.43     &     86.46   &  77.36     &    8.00    &  \textbf{1.22}    \\
$K=64$      &    \textbf{91.84}  &  \textbf{84.98}   &   \textbf{5.06} &  \textbf{1.38}      &    \textbf{86.86}  &  \textbf{77.80}   &   \textbf{7.62} &  {1.26}     \\
$K=128$       &    91.36    &   84.22    &   5.80     &   1.45    & 86.31       &  77.47     &   8.07     &  1.23    \\
$K=256$   &     91.27   &   84.07    &     5.41   &   1.50    &   86.14     &  77.16     &    7.91    &   1.23   \\
\bottomrule
\end{tabular}
\caption{Ablation study on channel selection count.}
  \label{tab:tab_AS_channelCount}
\end{table}

\begin{table}[t!]
  \centering
  \scriptsize
  \renewcommand{\arraystretch}{1.0}
  \setlength\tabcolsep{3.8pt}
\begin{tabular}{c|cccc|cccc}
 \toprule
\multirow{2}{*}{Settings} & \multicolumn{4}{c|}{Left Atrium~(20\%)} & \multicolumn{4}{c}{BraTS-2019~(20\%)} \\
\cline{2-9}
                          & Dice & Jaccard & 95HD & ASD   & Dice & Jaccard & 95HD & ASD  \\
\midrule
Random    &   90.67   &   83.11   &  6.03 &  1.60    &   85.88   &   76.92   &   8.17  &  1.30    \\
Router   &    \textbf{91.84}  &  \textbf{84.98}   &   \textbf{5.06} &  \textbf{1.38}      &    \textbf{86.86}  &  \textbf{77.80}   &   \textbf{7.62} &  \textbf{1.26}  \\
\bottomrule
\end{tabular}
\caption{Ablation study on the channel selection mechanism.}
\label{tab:tab_channel_select}
\end{table}

\textbf{Effect of Channel Selection Count.}
To further investigate the impact of channel selection ratio on model performance, we conduct an ablation study on the channel selection ratio, aiming to ensure that additional information is fully utilized while minimizing unnecessary interference. Note that we only modify the number of selected channels (\emph{i.e.}, Top-$K$), while the other strategies (\emph{i.e.}, CR and SSP) remain unchanged. 
As shown in Table~\ref{tab:tab_AS_channelCount}, full-channel interaction ($K=256$) yields the lowest performance, with a Dice score of $91.27\%$ on the LA dataset using $20\%$ labeled data. This degradation occurs because the full-channel approach introduces excessive noise and redundant information, which distracts the model from key features. While reducing $K$ to $32$ improves the Dice score to $91.66\%$, the gain remains limited. It can be observed that we can achieve the best performance when $K=64$.

\textbf{Effect of Channel Selection Mechanism.}
We further conduct an ablation study to investigate the impact of the channel selection mechanism on model performance, comparing the router method with the random channel selection method. This study aims to analyze the advantages of the router, which dynamically selects key features in a learnable manner, compared to the random method. Note that the ablation study is conducted based on $K=64$. Table~\ref{tab:tab_channel_select} shows the router mechanism outperforming random selection by $1.17\%$ (LA) and $0.98\%$ (BraTS-2019) in Dice score with $20\%$ labeled data. This performance gain confirms the router's utility in mitigating redundant information interference and improving model robustness.

\begin{figure}[t]
\centering
\includegraphics[width=0.99\columnwidth]{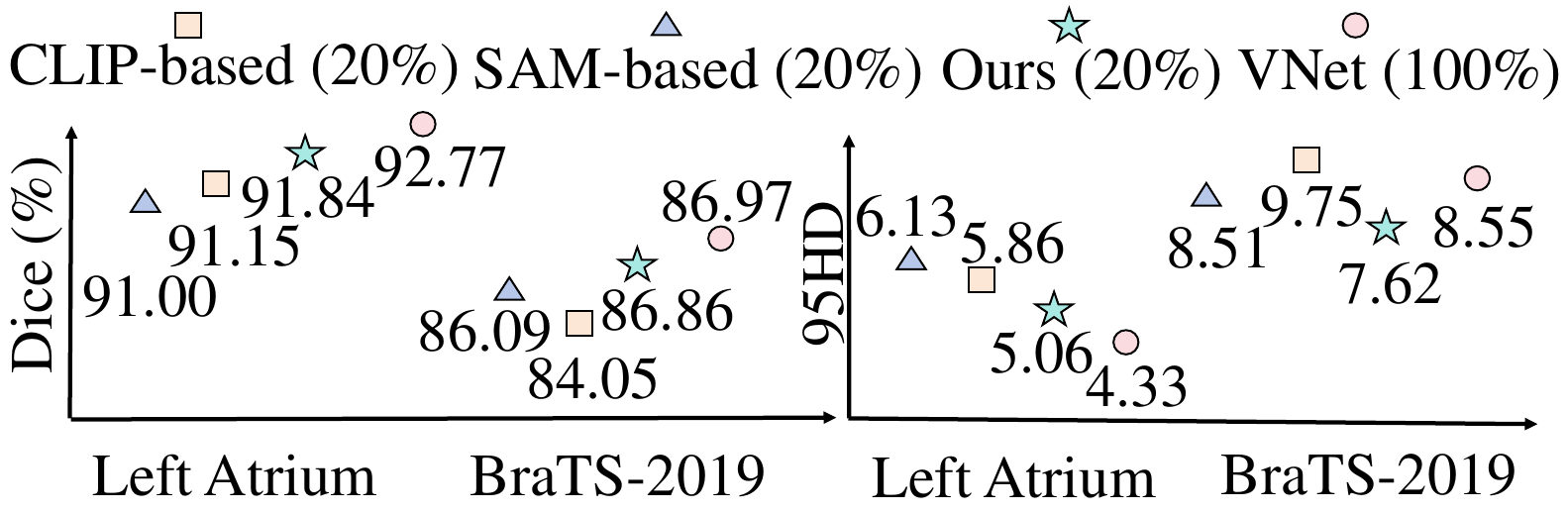}
\caption{Performance comparison of our method with foundation model-based semi-supervised approaches and the fully supervised VNet, using a $20\%$ labeled data ratio.
}
\label{fig4}
\end{figure}

\textbf{In-depth Comparison and Discussion.}
To more comprehensively evaluate the effectiveness of our method, we compare it with foundation model-based semi-supervised methods (VCLIP~\cite{li2024vclipseg} and SFR~\cite{li2025stitching}) and the fully-supervised VNet. The results in Fig.~\ref{fig4} demonstrate that our method achieves performance comparable to the fully-supervised paradigm. More significantly, on the BraTS-2019 dataset with only $20\%$ labeled data, our method surpasses the fully-supervised VNet in the 95HD metric, underscoring its exceptional precision.

\section{Conclusion}

We present BCSI, a novel semi-supervised paradigm that enhances interactions between labeled and unlabeled data. Our framework incorporates a Semantic-Spatial Perturbation (SSP) strategy, which enriches data diversity via strong augmentations while leveraging predictions from weak augmentations as pseudo-labels. Moreover, we propose a Channel-selective Router (CR) and a Bidirectional Channel-wise Interaction (BCI) module to dynamically select informative feature channels and enable bidirectional exchange between labeled and unlabeled data streams. Experimental results demonstrate that BCSI outperforms existing state-of-the-art semi-supervised medical image segmentation methods.



\bibliography{aaai2026}

\end{document}